\documentclass[letterpaper, 10 pt, conference]{ieeeconf}  

\IEEEoverridecommandlockouts                              

\usepackage{hyperref}       
\usepackage{url}            
\usepackage{booktabs}       
\usepackage{amsfonts}       
\usepackage{nicefrac}       
\usepackage{microtype}      
\usepackage{xcolor}         
\usepackage{cite}
\usepackage{bm}
\usepackage{mathtools}
\usepackage{multirow}
\usepackage{wrapfig}
\usepackage{graphicx}
\usepackage{subcaption}
\usepackage{caption} 
\usepackage{makecell}
\usepackage{cuted}   
\usepackage{capt-of} 

\title{\LARGE \bf
RAAP: Retrieval-Augmented Affordance Prediction with Cross-Image Action Alignment
}

\author{
Qiyuan Zhuang$^{1,2}$,
He-Yang Xu$^{1}$,
Yijun Wang$^{1}$,
Xin-Yang Zhao$^{3}$,
Yang-Yang Li$^{3}$,
Xiu-Shen Wei$^{1}$$^\dagger$
\thanks{This work was supported by National Natural Science Foundation of China under Grant (62522602), Basic Research Program of Jiangsu under Grant (BK20250073), CIE-Tencent Robotics X Rhino-Bird Focused Research Program, and the Fundamental Research Funds for the Central Universities (4009002401, 2242025K30024). This work was also supported by the Big Data Computing Center of Southeast University.}
\thanks{$^\dagger$ For Correspondence: {\tt\small weixs@seu.edu.cn}}
\thanks{$^{1}$School of Computer Science and Engineering, and Key Laboratory of New Generation Artificial Intelligence Technology and Its Interdisciplinary Applications, Southeast University, Nanjing 211189, China}
\thanks{$^{2}$Southeast University-Monash University Joint Graduate School, Southeast University, Suzhou 215123, China}
\thanks{$^{3}$School of Computer Science and Engineering, Nanjing University of Science and Technology, Nanjing 210094, China}
}

\begin{document}

\maketitle
\thispagestyle{empty}
\pagestyle{empty}


\begin{strip}
\vspace{-30mm}
\centering
\includegraphics[width=\textwidth]{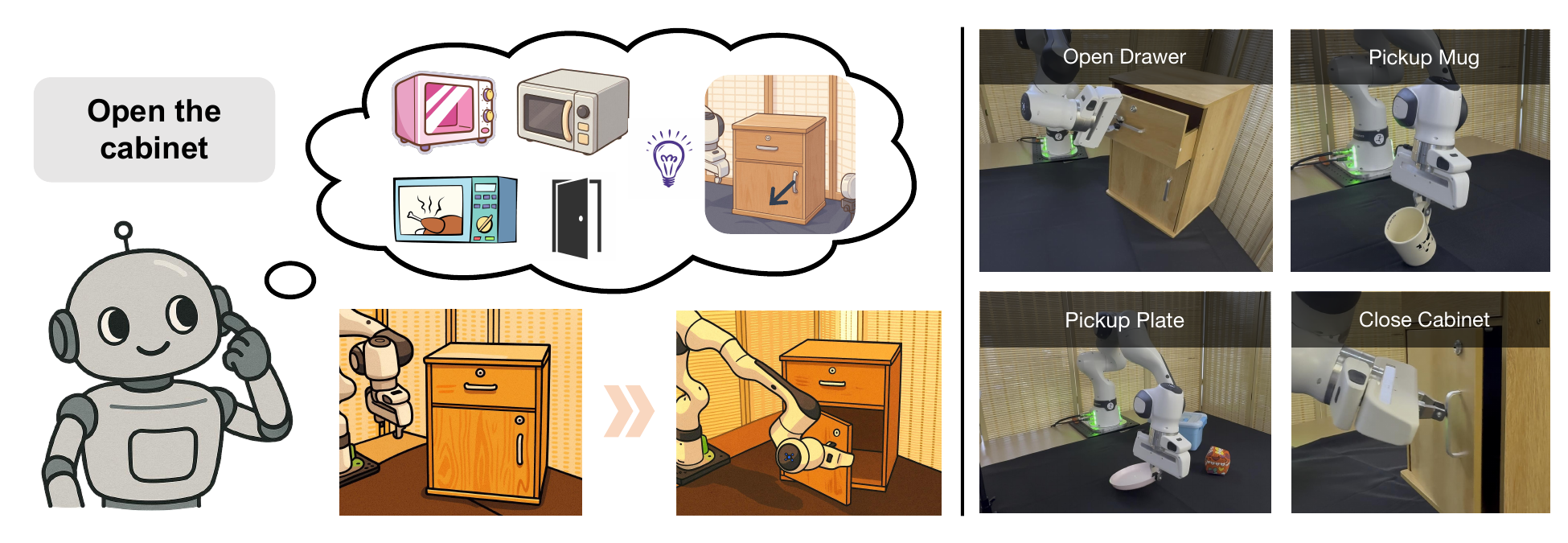}
\captionof{figure}{
When facing a novel task or unseen object category (e.g., ``open the cabinet''), 
\textbf{Retrieval-Augmented Affordance Prediction (RAAP)} retrieves semantically related experiences (e.g., ``opening a microwave'') and transfers the corresponding affordances to guide execution.
}
\label{fig:teaser}
\vspace{-1.2em}
\end{strip}

\begin{abstract}

Understanding object affordances is essential for enabling robots to perform purposeful and fine-grained interactions in diverse and unstructured environments. However, existing approaches either rely on retrieval, which is fragile due to sparsity and coverage gaps, or on large-scale models, which frequently mislocalize contact points and mispredict post-contact actions when applied to unseen categories, thereby hindering robust generalization. We introduce \textbf{Retrieval-Augmented Affordance Prediction (RAAP)}, a framework that unifies affordance retrieval with alignment-based learning. By decoupling static contact localization and dynamic action direction, RAAP transfers contact points via dense correspondence and predicts action directions through a retrieval-augmented alignment model that consolidates multiple references with dual-weighted attention. Trained on compact subsets of DROID and HOI4D with as few as tens of samples per task, RAAP achieves consistent performance across unseen objects and categories, and enables zero-shot robotic manipulation in both simulation and the real world. Project website: \href{https://github.com/SEU-VIPGroup/RAAP}{github.com/SEU-VIPGroup/RAAP}. 

\end{abstract}

\section{Introduction}

Object affordances provide the perceptual basis for robotic fine manipulation in diverse and unstructured environments~\cite{jamone2016affordances,yamanobe2017brief,hassanin2021visual, BCI}. Given a manipulation task, a robot must reason about \emph{where} to act (e.g., the graspable point on a drawer handle) and \emph{how} to act (e.g., pulling direction to open it) from raw visual observations. Such reasoning should extend beyond object categories to fine-grained object parts and motion cues: for instance, identifying the exact rim of a bowl to pick it up, or the precise orientation to lift a mug by its handle. While visual affordance learning has been widely studied in robotics and computer vision~\cite{do2018affordancenet,wu2020learning,mandikal2021learning}, scaling affordance prediction to new object instances and categories remains a persistent challenge, especially under limited robot data.

A growing line of work addresses this challenge from two distinct paradigms. The first is the \emph{retrieval-based paradigm}, where robots recall past demonstrations from a memory and transfer the associated affordances to new scenes~\cite{ju2024robo,kuang2024ram,wu2025afforddp}. This strategy has shown promising adaptability, as it leverages large collections of human and robot interaction data without requiring in-domain training, and can often generalize across tasks with minimal additional effort. However, retrieval methods face inherent limitations: the \emph{sparsity problem}, where reliance on a single top-1 match makes predictions fragile; and the \emph{coverage problem}, where the memory lacks semantically relevant instances, leading to failures on unseen categories.

The second is the \emph{training-based paradigm}, which learns predictive affordance models directly from large-scale data~\cite{bahl2023affordances,mo2021where2act,xu2025a0,tang2025uad,luo2022learning}. Such models can capture transferable visual patterns and achieve impressive generalization when abundant demonstrations are available. However, they also suffer from critical drawbacks. Many struggle to localize precise contact points or to predict reliable post-contact action directions~\cite{bahl2023affordances,mo2021where2act,xu2025a0}, and some restrict affordances to static contacts only without modeling the dynamic component~\cite{tang2025uad,luo2022learning}. As a result, training-based approaches alone remain insufficient for robust real-world generalization. 



In this paper, we present the \textbf{Retrieval-Augmented Affordance Prediction (RAAP)}, a framework that unifies the strengths of retrieval- and training-based approaches for task-aware affordance prediction. Our key insight is not merely that affordances can be represented as two components---a \emph{static contact point}, indicating where to act, and a \emph{dynamic action direction}, indicating how to act after contact---but that these components exhibit different uncertainty characteristics. While such representations have been explored in prior work~\cite{kuang2024ram}, existing approaches typically treat them as jointly transferable attributes. In contrast, we observe that static contact localization is primarily governed by geometric correspondence and can be reliably transferred from a single well-matched reference, whereas post-contact action direction is inherently more ambiguous and requires aggregating evidence across multiple references.

RAAP resolves these issues through a complementary inference design. For static affordance, we adopt dense feature correspondence with the top-1 retrieved reference to localize contact points~\cite{ju2024robo,kuang2024ram}. For dynamic affordance, we introduce a retrieval-augmented alignment model that aggregates cues from multiple references~\cite{rankrag, multidemo,pmlr-v267-han25d}. By conditioning visual features on action vectors and integrating them via a dual-weighted attention mechanism, the model selectively emphasizes task-relevant priors while suppressing noisy or misaligned examples. Crucially, by consolidating information across diverse exemplars, RAAP significantly reduces directional prediction errors and improves robustness under visual and geometric variations. This design enables RAAP to achieve strong performance with data-scarce settings (as few as tens of samples per task). Moreover, our method enables zero-shot robotic manipulation in both simulation and real-world environments. 

Our contributions are summarized as follows: 
\begin{itemize}
    \item We propose \textbf{RAAP}, a unified retrieval- and training-based paradigm that addresses the limitations of existing methods and enables generalization under data scarcity, achieving strong performance with only a handful of training samples per task.
    \item We design a novel \textbf{retrieval-augmented alignment model} that aggregates multiple references with dual-weighted attention, while treating static and dynamic affordances with complementary mechanisms.
    \item We conduct comprehensive evaluations on DROID, HOI4D, and real-world platforms, demonstrating that RAAP outperforms baselines (RAM, A0) in both \emph{unseen-object} and \emph{cross-category} generalization scenarios.
\end{itemize}

\begin{figure*}[t]
\vspace{0.5em}
  \centering
  \includegraphics[width=\textwidth,scale=1.00]{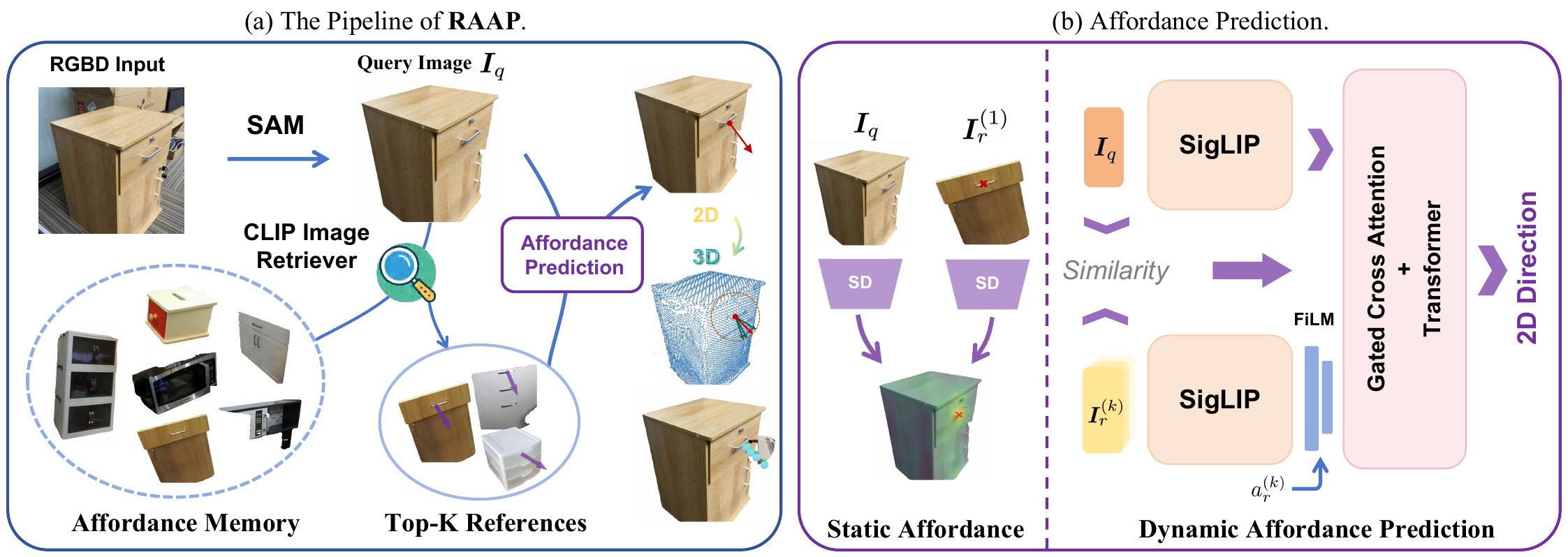}

     \caption{\textbf{Overview of the Retrieval-Augmented Affordance Prediction (RAAP) framework.} 
    (a) \textit{Pipeline.} Given an RGB-D input and a task label, RAAP retrieves top-$K$ references from an affordance memory using CLIP-based similarity. 
    It then predicts a 2D affordance (contact point and action direction) and lifts it to 3D for execution. 
    (b) \textit{2D Affordance Prediction.} 
    Static contact points are localized via dense correspondence using Stable Diffusion (SD) features, while dynamic action directions are inferred by a cross-image alignment module. 
    Both query and reference images are encoded with a shared SigLIP-2 backbone; reference tokens are further modulated by their action vectors via FiLM, and fused with query tokens through gated cross-attention and a Transformer.}
  
  \label{fig:pipeline}
  \vspace{-1.6em}
\end{figure*}

\section{Related Work}

\subsection{Visual Affordance Learning for Robotics}

Visual affordance learning~\cite{liu2022joint,yuan2024general,mo2021where2act,wu2022vatmart,wang2022adaafford,geng2022end} aims to infer \emph{where} and \emph{how} an object can be interacted with from sensory inputs, providing crucial perceptual cues for downstream planning and control. Some prior works have employed pixel-level segmentation or detection to infer affordance regions~\cite{do2018affordancenet,myers2015affordance}. While intuitive, such heatmaps are often diffuse and lack the geometric precision required for complex manipulation. Subsequent work therefore explores richer representations, including dense visual correspondence~\cite{hadjivelichkov2023one} and per-point affordance prediction on 3D point clouds~\cite{qin2023dexpoint}. For articulated objects, keypoint-based affordances provide a compact abstraction of interaction~\cite{manuelli2019kpam,affkp}. 
In parallel, UAD~\cite{tang2025uad} distills affordance knowledge from foundation models to predict static contact maps from single images.

However, static affordances alone cannot capture the dynamic nature of interactions. 
Recent work therefore models dynamic affordances that encode both contact and post-contact motion, such as VRB~\cite{bahl2023affordances} and A0~\cite{xu2025a0}. 
Instead of relying solely on large curated datasets, our work combines retrieval from a compact affordance memory with alignment-based learning to enable generalization under limited data.

\subsection{Zero-Shot Robotic Manipulation}

Zero-shot robotic manipulation~\cite{sontakke2023roboclip,huang2024copa,ding2024open6dor,li2024ag2manip,fangandliu2024moka,bharadhwaj2024towards} aims to enable robots to perform new tasks from high-level instructions without task-specific training. Existing approaches mainly follow two paradigms: large-scale imitation learning and retrieval-based methods. 
Large-scale imitation learning trains generalizable policies from diverse robot and human demonstrations~\cite{black2025pi0,kim24openvla}, often leveraging multimodal foundation models for semantic grounding~\cite{Yin_2024, fgmllm, dlllm, shen2025exp,yu2025benchmarking}, but remains heavily dependent on data scale and diversity. 

Retrieval, as a general mechanism for leveraging prior instances across domains~\cite{fgia-survey,suo2024knowledge,gordo2016deep,multiuselearning,wei2023attr}, has recently been adopted for zero-shot robotic manipulation. Some leverage semantic correspondence to transfer affordances across object categories~\cite{ju2024robo,zhu2024densematcher}. RAM~\cite{kuang2024ram} builds a cross-domain interaction memory for affordance transfer, while AffordDP~\cite{wu2025afforddp} integrates retrieval with diffusion policies to transfer 3D affordances through semantic matching and geometric alignment. 
Our method follows this paradigm but introduces retrieval-augmented alignment that consolidates multiple references and explicitly decouples static and dynamic affordances, enabling robust zero-shot manipulation under limited data.

\section{Method}

In this section, we describe (A) how we construct an affordance memory from prior interactions, (B) how we retrieve task-relevant references and transfer static contact points, (C) how we predict dynamic action directions via retrieval-augmented alignment, and (D) how we lift 2D affordances into 3D for robotic execution. As illustrated in Fig.~\ref{fig:pipeline}, RAAP decomposes affordance into static (contact point) and dynamic (action direction) components, which are predicted through complementary retrieval and alignment.

\subsection{Affordance Memory}

We represent a 2D affordance as a tuple $\mathcal{A}^{\text{2D}} = (\bm{c}^{\text{2D}}, \bm{a}^{\text{2D}})$, where $\bm{c}^{\text{2D}}, \bm{a}^{\text{2D}} \in \mathbb{R}^2$ denote the contact point (static affordance) and the post-contact motion direction (dynamic affordance), respectively. Both quantities are defined in the image coordinate frame. Given a target image $\bm{I} \in \mathbb{R}^{H \times W \times 3}$, the goal is to estimate the affordance that best facilitates task execution in the current context. At execution time, the 2D affordance can be lifted to 3D using object point clouds and camera intrinsics (see Sec.~\ref{sec:execution}).

To support retrieval-based generalization, we construct a visual affordance memory $\mathcal{R}$ that stores prior segmented object appearances and their associated interactions. We begin by applying Grounded-SAM~\cite{ren2024grounded} to segment the target object from the source image $\bm{I}$, producing a cropped image $\bm{I}_g$. Using a CLIP~\cite{radford2021learning} image encoder, we extract an appearance embedding $f$ from $\bm{I}_g$ for retrieval. Each memory entry stores a tuple of the form:
\begin{equation}
    \mathcal{R} = \{(I_g, f, T, \mathcal{A}^{\text{2D}})\},
\end{equation}
where $T$ is the task label, and $\mathcal{A}^{\text{2D}}$ denotes the 2D affordance annotation. 

For dynamic affordance extraction, we convert annotated 2D trajectories into normalized motion directions. Specifically, each trajectory is reduced to its dominant orientation, which is then represented as a unit vector $\bm{a}^{\text{2D}}$. This ensures that the dynamic component encodes only the intended action direction, independent of trajectory length or scale.

The memory is populated from a composite dataset containing DROID and HOI4D. We filter out samples whose trajectories do not yield valid post-contact directions.

\subsection{Retrieval and Static Affordance Transfer}
\label{sec:static_affordance}
Given a new target scene, we perform retrieval from the affordance memory in two stages. First, we use a CLIP text encoder~\cite{radford2021learning} to filter memory entries by task relevance, identifying those whose task labels match or are semantically similar to the current task $T$. Within this task-constrained subset, we compute cosine similarity between the CLIP embeddings of the query image $\bm{I}_q$ and the stored exemplars $\{\bm{I}_r^{(k)}\}$, and retrieve the top-$K$ most similar entries. These retrieved examples will be used differently for static and dynamic affordance prediction.

To localize the static contact point $\bm{c}_q^{\text{2D}}$ in the query image, we adopt a dense feature matching strategy based on the top-1 retrieved example. Let $\bm{I}_r^{(1)}$ denote the top-1 retrieved reference image and $\bm{c}_r^{\text{2D}}$ its annotated contact point. We extract dense per-pixel visual features using a Stable Diffusion (SD)~\cite{rombach2022high} encoder and upsample them to the original resolution. To estimate $\bm{c}_q^{\text{2D}}$, we compare the local feature around $\bm{c}_r^{\text{2D}}$ against all pixel features in $\bm{I}_q$ by identifying the location with maximal cosine similarity. This procedure enables direct transfer of static affordance through high-resolution correspondence in feature space, preserving fine spatial structures that are critical for accurate contact localization.

We emphasize that our framework treats static and dynamic affordance components separately: the contact point $\bm{c}_q^{\text{2D}}$ is estimated via one-shot dense matching from the most similar retrieved example (top-1), as SD-based features already provide reliable correspondences even under appearance variations, while the direction vector $\bm{a}_q^{\text{2D}}$ is predicted via a retrieval-augmented alignment model that consolidates directional priors across multiple references (see Sec.~\ref{sec:retrieval_modeling}).

\subsection{Learning Dynamic Affordance via Retrieval-Augmented Alignment}
\label{sec:retrieval_modeling}

Unlike static contact localization, dynamic action prediction is more abstract and often fails with a single reference. We therefore aggregate top-$K$ exemplars to reduce directional ambiguity and mitigate errors from mis-retrieval. 
To predict the post-contact action direction $\bm{a}_q^{\text{2D}}$, we introduce a retrieval-augmented cross-image action alignment module that conditions on multiple prior interaction examples stored in the affordance memory. The model learns to align cross-scene visual cues and manipulation intents through token-level attention and dynamic weighting.

Given a query image $\bm{I}_q$ and a task label $T$, we first retrieve the top-$K$ most visually similar entries ${(\bm{I}_r^{(k)}, \mathcal{A}_r^{(k)}, s^{(k)})}_{k=1}^{K}$ from the task-relevant subset of memory $\mathcal{R}$. Each entry contains a reference image $\bm{I}_r^{(k)}$, similarity score $s^{(k)}$, and annotated 2D affordance $\mathcal{A}_r^{(k)} = (\bm{c}_r^{(k)}, \bm{a}_r^{(k)})$. In this stage, only the directional component $\bm{a}_r^{(k)}$ is used, as the contact point $\bm{c}_q^{\text{2D}}$ has already been estimated.

\noindent\textbf{Action-Conditioned Reference Encoding} 
To condition each reference image on its associated manipulation intent, we incorporate action vectors into the patch-level feature representation extracted by a shared SigLIP-2 encoder~\cite{tschannen2025siglip}. For each retrieved reference image $\bm{I}_r^{(k)}$, we obtain a sequence of patch tokens $\mathbf{F}_r^{(k)} \in \mathbb{R}^{N \times d}$, where $N$ is the number of patches and $d$ is the feature dimension.

The associated 2D action vector $\bm{a}^{(k)} \in \mathbb{R}^2$ is projected via a multi-layer perceptron (MLP) into modulation parameters $\gamma(\bm{a}^{(k)}), \beta(\bm{a}^{(k)}) \in \mathbb{R}^d$. We then apply FiLM-style modulation~\cite{perez2018film} to each patch feature independently:
\begin{equation}
\tilde{\mathbf{F}}_r^{(k)}[i] = \gamma(\bm{a}^{(k)}) \cdot \mathbf{F}_r^{(k)}[i] + \beta(\bm{a}^{(k)}), \quad \text{for } i = 1,\dots,N.
\end{equation}

This yields action-conditioned features $\tilde{\mathbf{F}}_r^{(k)} \in \mathbb{R}^{N \times d}$ that jointly encode visual context and motion intent. To distinguish different references, we add a learnable reference ID embedding $\mathbf{E}_{\text{ref}}^{(k)} \in \mathbb{R}^{1 \times d}$:
\begin{equation}
\hat{\mathbf{F}}_r^{(k)} = \tilde{\mathbf{F}}_r^{(k)} + \mathbf{E}_{\text{ref}}^{(k)}.
\end{equation}
The resulting tokens $\hat{\mathbf{F}}_r^{(k)}$ are used as memory inputs for attention-based alignment.

\noindent\textbf{Cross-Attention Aggregation} 
We apply a cross-attention module to fuse the action-conditioned reference tokens into the target representation, followed by a Transformer encoder~\cite{vaswani2017attention} that refines the fused target representation after retrieval-conditioned alignment. The target image tokens $\mathbf{F}_q \in \mathbb{R}^{N_q \times d}$ serve as queries, while the action-conditioned reference tokens $\{\hat{\mathbf{F}}_r^{(k)}\}_{k=1}^K$ are concatenated along the sequence dimension, forming a unified key-value matrix $\mathbf{F}_{\text{mem}} \in \mathbb{R}^{K \cdot N \times d}$. This cross-attention mechanism enables the target representation to selectively incorporate directional cues from multiple retrieved exemplars.

To address the varying relevance of retrieved samples, we introduce a dual-weighting strategy that combines learned semantic relevance with pre-computed visual similarity. First, we compute global semantic features by averaging patch-level tokens:
\begin{align}
\mathbf{z}_q &= \frac{1}{N_q} \sum_{i=1}^{N_q} \mathbf{F}_q[i], \label{eq:zq} \\
\mathbf{z}_r^{(k)} &= \frac{1}{N} \sum_{i=1}^{N} \hat{\mathbf{F}}_r^{(k)}[i], \label{eq:zr}
\end{align}
where $\mathbf{z}_q, \mathbf{z}_r^{(k)} \in \mathbb{R}^d$ represent the global semantic representations of the target and $k$-th reference image respectively.
We compute semantic relevance weights using a lightweight gating network $w_k = \sigma(\text{MLP}([\mathbf{z}_q; \mathbf{z}_r^{(k)}]))$, where $\sigma$ is the sigmoid activation and $[\cdot;\cdot]$ denotes feature concatenation.

The final attention weights $\{w_k^{\text{final}}\}_{k=1}^K$ are computed by combining the learned gating weights with the pre-computed similarity scores $\{s^{(k)}\}_{k=1}^K$:
\begin{equation}
w_k^{\text{final}} = \frac{\text{softmax}(s^{(k)}) \cdot w_k}{\sum_{j=1}^K \text{softmax}(s^{(j)}) \cdot w_j + \epsilon},
\end{equation}
where $\epsilon > 0$ is a small constant for numerical stability.

The fused target tokens are concatenated with a learnable [CLS] token and passed through a Transformer encoder. The final action direction $\bm{a}_q^{\text{2D}}$ is regressed from the [CLS] token via a lightweight MLP.

\noindent\textbf{Learning Objective} 
Ground-truth action directions $\hat{a}_q^{\text{2D}}$ are stored as unit vectors. We train the model by regressing the $(x,y)$ components using a mean squared error (MSE) loss:
\begin{equation}
\mathcal{L} = \tfrac{1}{2}\,\| \bm{a}_q^{\text{2D}} - \bm{\hat{a}}_q^{\text{2D}} \|_2^2.
\end{equation}
At inference time, the predicted vector $\bm{a}_q^{\text{2D}}$ is $\ell_2$-normalized before evaluation.

\subsection{Execution via Sampling-Based Affordance Lifting}
\label{sec:execution}

To execute the predicted 2D affordance $(c_q^{\text{2D}}, a_q^{\text{2D}})$, we first lift it into the 3D workspace using camera intrinsics and the observed point cloud. The 2D contact point $c_q^{\text{2D}}$ is projected to a set of candidate 3D points, from which we select the closest valid surface point $\bm{c}_q^{\text{3D}}$ as the target grasp region.

We then apply a dense grasp sampler~\cite{wang2021graspness} on the cropped point cloud around $\bm{c}_q^{\text{3D}}$ to generate a set of grasp proposals. The grasp pose closest to $\bm{c}_q^{\text{3D}}$ is selected for execution. To reach this selected grasp pose, we solve inverse kinematics (IK) to generate a collision-free joint-space trajectory, ensuring precise and feasible grasp approach planning.

After grasping, the 2D direction $\bm{a}_q^{\text{2D}}$ is transformed into a 3D displacement vector $\bm{\tau}_q^{\text{3D}}$ using the local surface orientation and camera geometry. The robot then executes the post-contact action by moving the end-effector along $\bm{\tau}_q^{\text{3D}}$. In simulation, we use position control for simplicity. In the real world, we adopt real-time Cartesian impedance control to modulate interaction forces during execution, allowing for compliant and safe manipulation.

\section{Experiments}

\subsection{Task-Aware Affordance Prediction}
\label{sec:exp_taskaware}

We first evaluate RAAP on task-aware dynamic affordance prediction, focusing on the post-contact action direction. 
Static affordance (i.e., contact point) is excluded here, as dense feature-based transfer has been extensively validated in prior work~\cite{ju2024robo, wu2025afforddp, kuang2024ram}, 
and will instead be further assessed in our real-world experiments.

\noindent\textbf{Datasets} We experiment on subsets of the DROID~\cite{khazatsky2024droid} and HOI4D~\cite{liu2022hoi4d} datasets, adopting a 70/30 train–test split per task. 
To avoid potential data leakage, we remove samples with nearly identical object instances and viewpoints. 
For RAAP with $K>0$, we further construct a task-aware affordance memory by merging semantically related tasks (e.g., \emph{open drawer} and \emph{open microwave}), which enables cross-task transfer of interaction cues during retrieval. 
On average, the DROID subsets provide about 18 samples per task, while HOI4D provides about 160. We use the same datasets as RAM~\cite{kuang2024ram}, but apply additional cleaning to ensure fairer train–test separation.

\noindent\textbf{Evaluation Metric} Given a predicted unit vector $\bm{a}_q^{\text{2D}}$ and ground-truth $\bm{\hat a}_q^{\text{2D}}$, we report the Mean Angular Error (MAE):
\begin{equation}
\text{MAE} = \arccos\!\big( \langle \bm{a}_q^{\text{2D}}, \bm{\hat a}_q^{\text{2D}} \rangle \big) \cdot 180 / \pi,
\end{equation}
which directly measures angular discrepancy in degrees. Compared to vector MSE, MAE is invariant to magnitude and provides more interpretable directional accuracy.

\noindent\textbf{Experimental Setting} 
RAAP employs a 6-layer transformer with SigLIP-2 (base, patch16, 384 resolution) as the visual backbone, and all parameters are updated during training. 
For $K>0$, each query retrieves 10–20 candidates from memory, from which $K$ references are sampled per episode. 
To enhance training diversity, this sampling is repeated five times per query. 
We also apply data augmentation by horizontally flipping both the input images and the associated action directions. 
Models are trained for up to 50 epochs with early stopping if the training loss does not improve for 5 consecutive epochs. 
On a single NVIDIA RTX 4090 GPU, training one task of the DROID dataset requires approximately 85 seconds.
All reported results are averaged over three runs with different random train–test splits to reduce variance.

\begin{table}[t]
\vspace{0.5em}
\centering
\caption{MAE~$\downarrow$ for task-aware dynamic affordance prediction. 
The first six rows correspond to individual manipulation tasks from the DROID dataset, the seventh row reports the averaged performance on HOI4D pickup tasks (\emph{bowl}, \emph{bottle}, \emph{mug}). The final row shows the overall average across all tasks.}
\label{tab:direction}
\begin{tabular}{lcccc}
\toprule
Tasks & RAM & A0-170M & RAAP$_{K=0}$ & RAAP$_{K=3}$ \\
\midrule
Open microwave   &  49.82 & 119.64  &  51.75 &  \textbf{37.00} \\
Close microwave  &  55.74 &  103.10 &  61.38 &  \textbf{30.88} \\
Open drawer      &  51.73 &  112.89 &  \textbf{15.45} &  16.23 \\
Close drawer     &  83.04 &  85.16 &  52.11 &  \textbf{48.25} \\
Pickup bowl      &  53.85 &  34.37 &  36.76 &  \textbf{31.45}  \\
Pickup bottle   & 82.62 &  \textbf{33.86}  & 38.70  &  37.67  \\
Pickup [obj] (HOI) &  63.07   &  34.63   &   26.97  &  \textbf{26.36}  \\
\midrule
Overall avg.     &  62.84 &  74.81 &  40.45 &  \textbf{32.55} \\
\bottomrule
\end{tabular}

\vspace{-1em}
\end{table}

\begin{figure}[t]  
    \centering
    \includegraphics[width=\linewidth]{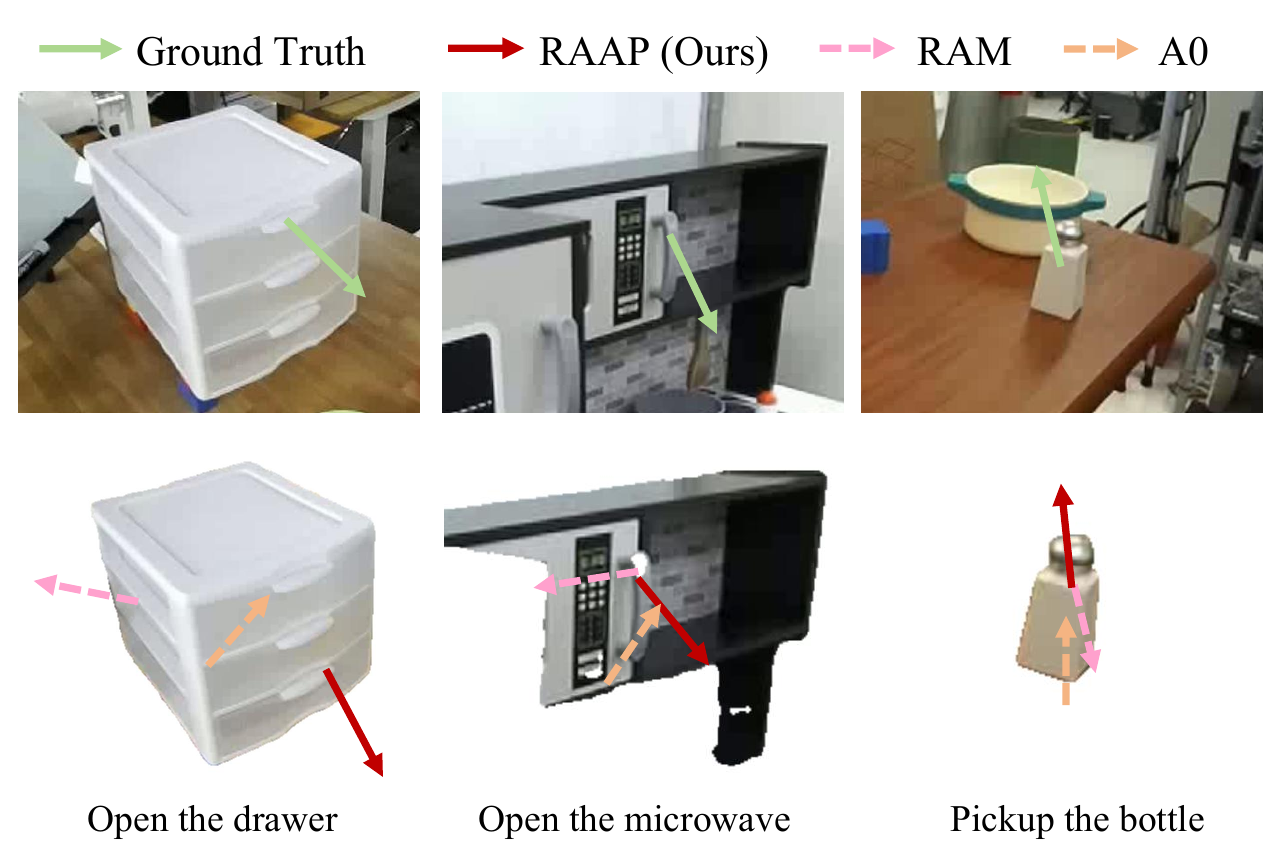}
    \caption{Qualitative comparison of 2D affordance predictions on the DROID dataset. 
The first row shows the input RGB image with ground-truth affordances (contact point and action direction), and the second row visualizes predictions from RAM, A0, and RAAP ($K=3$). 
    }
    \label{fig:comparison}
    \vspace{-1.8em}
\end{figure}

\noindent\textbf{Baselines} 
We compare against two representative methods: RAM~\cite{kuang2024ram}, a retrieval-and-transfer framework without multi-reference alignment, and A0~\cite{xu2025a0}, a large-scale affordance-aware diffusion model trained on DROID, HOI4D, and ManiSkill (we report the A0-170M variant). 
Note that for A0, we only use the first two predicted 2D trajectory points to extract the contact location and action direction.

\noindent\textbf{Main Results} 
Figure~\ref{fig:comparison} and Table~\ref{tab:direction} jointly present qualitative and quantitative comparisons. 
RAAP yields action directions that closely align with the ground truth across tasks, benefiting from the aggregation of multiple retrieved references rather than relying on a single exemplar. 
In contrast, RAM transfers features from only one reference, which often results in globally incorrect motion directions when the retrieved instance is not well aligned. 
A0 demonstrates reasonable generalization in \emph{pickup} actions, but its predictions for \emph{open}/\emph{close} tasks deviate substantially from the ground truth in both direction and contact localization. 

Table~\ref{tab:direction} further compares RAAP with RAM, A0, and our ablations using different numbers of retrieved references ($K=0$ vs. $K=3$). 
RAAP with retrieval ($K=3$) achieves the lowest average error of $32.55^\circ$, representing a reduction of nearly $50\%$ relative to RAM ($62.84^\circ$) and over $50\%$ relative to A0 ($74.8^\circ$), while also surpassing the no-retrieval variant ($K=0$).
The benefit of retrieval is particularly evident on the \emph{open}/\emph{close} tasks in DROID, where RAM and A0 often fail to predict meaningful directions. 
On HOI4D pickup tasks, RAAP also maintains the best overall accuracy. 
These results demonstrate that retrieval-augmented alignment substantially enhances dynamic affordance prediction, especially for semantically complex manipulations.

RAAP requires approximately 4~seconds per inference on an RTX~4090 GPU, 
which is substantially faster than RAM (50~seconds per inference) but slower than A0 (0.2~seconds).

\noindent\textbf{Ablation Studies}
Table~\ref{tab:dual_weight_ablation} reports an ablation on the proposed dual-weighted attention module, comparing the full model against several variants. 
Removing either gating (``w/o Gating'') or similarity weights (``w/o Similarity'') leads to marked performance drops, confirming that the two weighting signals are complementary: similarity provides a strong prior on appearance, while gating suppresses misaligned references. 
Uniform weighting (averaging all retrieved references with equal weights) performs better than either single-weight variant, but remains inferior to the full model. 
RAAP (Full) achieves the lowest errors, highlighting the importance of jointly leveraging both gating and similarity cues for robust dynamic affordance prediction.

\begin{figure}[t]  
    \centering
    \includegraphics[width=0.9\linewidth]{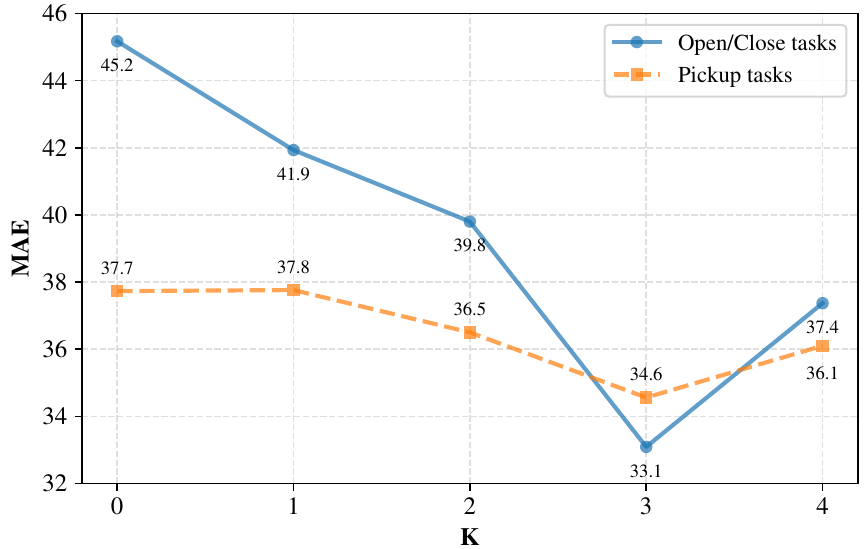}
    \caption{MAE~$\downarrow$ of RAAP as the number of retrieved references $K$ varies from 0 to 4 on the DROID dataset. 
    }
    \label{fig:K}
    \vspace{-1.6em}
\end{figure}

We further examine the effect of the number of retrieved references $K$ on dynamic affordance prediction using the DROID dataset. 
As shown in Fig.~\ref{fig:K}, on open/close tasks, RAAP achieves a substantial gain when increasing $K$ from 0 to 3, reducing the error from 45.2$^\circ$ to 33.1$^\circ$. 
A similar but milder trend is observed on pickup tasks. 
Notably, $K=3$ consistently yields the best performance across both task types, whereas too few references (e.g., $K=0$ or $K=1$) leave the model vulnerable to mis-retrieval, and too many (e.g., $K=4$) introduce noisy exemplars that slightly degrade accuracy. 
These results validate our design choice of aggregating a moderate number of references for robust dynamic prediction.

\begin{table}[t]
\vspace{0.5em}
\centering
\caption{Ablation on dual-weighted attention for dynamic affordance prediction (MAE~$\downarrow$, degrees). 
Results are reported on the DROID dataset for two task categories: \emph{open/close} (e.g., microwave) and \emph{pickup} (e.g., bowl, bottle).}
\begin{tabular}{lcc}
\toprule
Variant              & Open/Close & Pickup \\
\midrule
RAAP (Full)          & \textbf{33.09} & \textbf{34.56} \\
w/o Gating           & 36.14          & 42.92 \\
w/o Similarity       & 39.47          & 38.96 \\
w/ Uniform Weights   & 37.63          & 36.54 \\
\bottomrule
\end{tabular}
\label{tab:dual_weight_ablation}
\vspace{-0.5em}
\end{table}

\subsection{Simulation and Real World Experiments}
\label{sec:real}

\begin{table}[t]
\vspace{0.5em}
\centering
\footnotesize
\renewcommand{\arraystretch}{1.1}
\caption{Real-world manipulation success rates (20 trials per task). 
The top block shows \emph{unseen-object} generalization, and the bottom block shows \emph{cross-category} generalization. }
\label{tab:realworldacc}
\begin{tabular}{llccc}
\toprule
Settings & Tasks & RAM & A0 & RAAP~(Ours) \\
\midrule
\multirow{5}{*}{\makecell{Unseen \\ Objects}}
& Open drawer & 70\% & 0\% & \textbf{85\%}\\
& Close drawer & 60\% & 5\% & \textbf{80\%} \\
& Pickup bowl & 85\% & 65\% & \textbf{90\%} \\
& Pickup bottle & 50\% & 35\% &  \textbf{75\%} \\
& Pickup mug &  \textbf{70\%}  & 25\% & \textbf{70\%} \\
\midrule
\multirow{4}{*}{\makecell{Unseen \\ Categories}} 
& Open cabinet & 70\% & 0\% & \textbf{80\%} \\
& Close cabinet & 75\% & 15\% & \textbf{100\%} \\
& Pickup plate &  80\% & 20\%  & \textbf{85\%} \\
& Pickup watering can  & 50\% & 10\% & \textbf{60\%} \\
\bottomrule
\end{tabular}
\vspace{-1.6em}
\end{table}


\noindent\textbf{Experimental Setting} 
Real-world experiments were conducted on a Franka Emika Research~3 robot equipped with its native parallel-jaw gripper. 
Perception was provided by RGB-D input from either an Intel RealSense D455 (front view) or an Intel RealSense L515 (side view), with only one camera used per task and no multi-view fusion. 
All models were trained exclusively on subsets of the DROID and HOI4D datasets without using real-world demonstrations. 
Unless otherwise specified, RAAP and the baseline methods (RAM and A0) follow the same training and inference configurations as in Sec.~\ref{sec:exp_taskaware}, with RAAP using retrieval with $K=3$ by default. 
In both real-world and simulation studies, control follows the execution protocol described in Sec.~\ref{sec:execution}.

We evaluate two generalization scenarios:  
(i)~\emph{unseen-object} generalization, where the task (e.g., \emph{open drawer}) remains the same but the manipulated object instance differs from those observed during training; and  
(ii)~\emph{cross-category} generalization, where affordances are transferred to semantically related object categories (e.g., training only on \emph{open microwave} but testing on \emph{open cabinet}). 
Performance is measured by the success rate (SR), averaged over 20 trials per task with randomized object placements.

\noindent\textbf{Real-World Results} 
The results in Table~\ref{tab:realworldacc} reveal clear differences across methods. 
A0 consistently fails across both \emph{pickup} and \emph{open/close} tasks due to poor contact point prediction, which prevents stable grasping and reliable execution. 
RAM achieves moderate success, but its predicted action directions are often misaligned with the intended motion, leading to unstable or incomplete executions. 

RAAP demonstrates the most reliable performance across both generalization settings. 
In \emph{unseen-object} tasks, it improves over RAM by 15–25 points on \emph{open/close drawer} and achieves the highest success rates in all pickup tasks. For \emph{cross-category} generalization, RAAP attains 100\% success on \emph{close cabinet}. This task is relatively less sensitive to precise contact localization since the gripper can push on the cabinet door without needing to grasp the handle, and approximate directional accuracy is sufficient. 
In contrast, \emph{close drawer} requires accurate directional prediction to overcome the drawer’s resistance; errors in action direction lead to frequent failures for RAM and A0. 
RAAP, by leveraging retrieval-augmented cues, maintains high success despite these physical challenges. 
We observe occasional failures in pickup tasks (e.g., \emph{pickup mug}, 70\%), primarily due to the difficulty of grasping small object parts such as handles, even when the predicted 2D affordances are correct. 
As shown in Fig.~\ref{fig:teaser}, RAAP successfully handles both unseen-object and cross-category scenarios. 


\noindent\textbf{Simulation Results} 
We conduct a controlled study in MuJoCo~\cite{todorov2012mujoco} with a UR5e manipulator, following the same control protocol as in real-world experiments. 
To test cross-category generalization, RAAP is trained only on the \emph{pickup mug} task of HOI4D and evaluated on \emph{pickup kettle}, with the kettle placed at different positions within the workspace. 

\begin{figure}[t]
\vspace{0.5em}
    \centering
    \includegraphics[width=\linewidth]{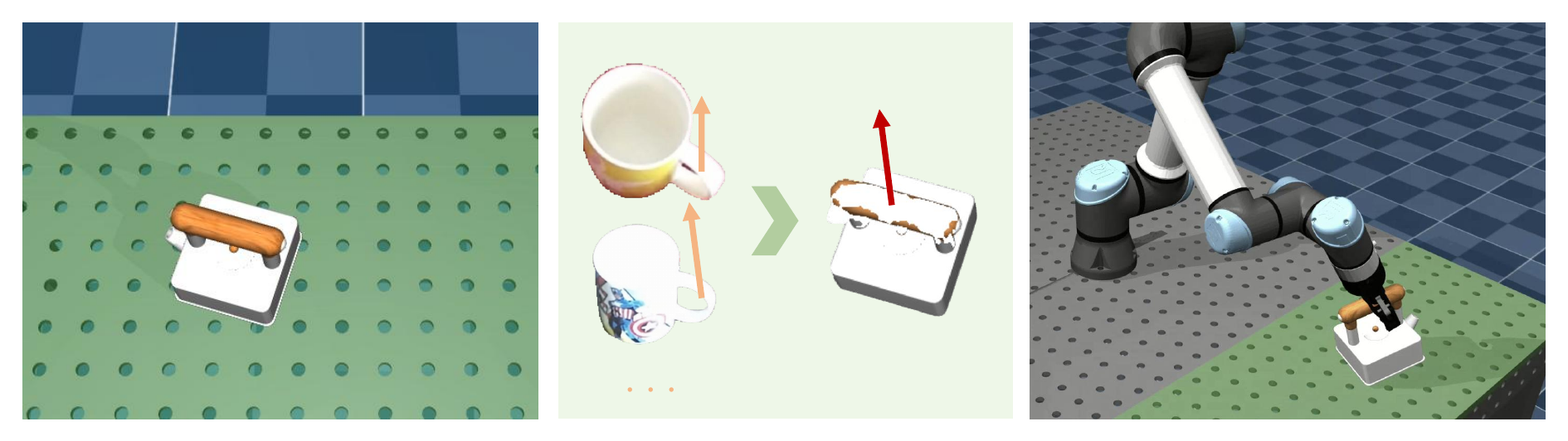} 
    \caption{
Qualitative results on \emph{pickup kettle} in MuJoCo with a UR5e manipulator. RAAP successfully transfers handle-oriented affordances to kettles. 
    }
    \label{fig:sim_kettle}

    \vspace{-1.8em}

\end{figure}

Over 20 trials, RAAP ($K=3$) achieves an SR of \textbf{85\%}, outperforming RAM (70\%) and A0 (10\%). 
Qualitatively (Fig.~\ref{fig:sim_kettle}), RAAP transfers handle-oriented action cues from mugs to kettles, enabling stable approach-and-lift behaviors. 
RAM, limited to single-exemplar transfer, frequently misaligns the predicted direction when the retrieved instance is not geometrically well matched, while A0 fails to localize valid graspable regions. 
This simulation study complements the real-world results, showing that RAAP provides robust cross-category transfer under geometric variations.

\section{Conclusion}

We presented the \textbf{Retrieval-Augmented Affordance Prediction (RAAP)}, a framework that unifies retrieval- and training-based approaches for task-aware affordance prediction. By decoupling static contact localization and dynamic action direction, RAAP leverages dense correspondence for pixel-level transfer while employing a retrieval-augmented alignment model to aggregate multiple references with dual-weighted attention. Our experiments on DROID, HOI4D, and real-world robotic platforms show that RAAP substantially improves generalization to unseen objects and novel categories, achieves competitive performance with only a handful of training samples per task, and enables zero-shot robotic manipulation in both simulation and the real world. These results indicate that RAAP provides a data-efficient and generalizable solution for fine-grained robotic manipulation.

Despite these promising results, several limitations remain. 
First, RAAP currently focuses on short-horizon tasks; extending it to multi-object and long-horizon sequential interactions would further test its robustness. 
Second, RAAP operates in an open-loop setting, and coupling it with closed-loop control policies could further improve robustness in dynamic and unstructured environments.

\addtolength{\textheight}{0cm}   











\bibliographystyle{IEEEtran}
\bibliography{root}

\end{document}